\documentclass[letterpaper]{article} 
\usepackage{aaai2026}  
\usepackage{times}  
\usepackage{helvet}  
\usepackage{courier}  
\usepackage[hyphens]{url}  
\usepackage{graphicx} 
\usepackage{natbib}  
\usepackage{caption} 
\usepackage{algorithm}
\usepackage{algorithmic}

\usepackage{newfloat}
\usepackage{listings}
\DeclareCaptionStyle{ruled}{labelfont=normalfont,labelsep=colon,strut=off} 
\floatstyle{ruled}
\newfloat{listing}{tb}{lst}{}
\floatname{listing}{Listing}
\usepackage{mathtools}
\usepackage{xspace}
\usepackage{xcolor}
\usepackage{amsfonts}
\usepackage{amsmath}
\usepackage{booktabs}
\usepackage{multirow}
\usepackage{amssymb}
\usepackage{array}
\usepackage{enumitem}
\makeatletter
\DeclareRobustCommand\onedot{\futurelet\@let@token\@onedot}
\def\@onedot{\ifx\@let@token.\else.\null\fi\xspace}

\def\eg{\emph{e.g}\onedot} 
\def\ie{\emph{i.e}\onedot}

\makeatother

\newcommand{\textbfit}[1]{\textbf{\emph{#1}}}

\title{Vision-Only Gaussian Splatting for Collaborative Semantic Occupancy Prediction}
\author{
  Cheng Chen\textsuperscript{\rm 1},
  Hao Huang\textsuperscript{\rm 2}\thanks{Corresponding authors.},
  Saurabh Bagchi\textsuperscript{\rm 1}\footnotemark[1]
}
\affiliations {
    \textsuperscript{\rm 1}Purdue University\\
    \textsuperscript{\rm 2}New York University Abu Dhabi\\
    chen4384@purdue.edu, hh1811@nyu.edu, sbagchi@purdue.edu
}

\usepackage{bibentry}

\begin{document}

\maketitle

\begin{abstract}
Collaborative perception enables connected vehicles to share information, overcoming occlusions and extending the limited sensing range inherent in single-agent (non-collaborative) systems. Existing vision-only methods for 3D semantic occupancy prediction commonly rely on dense 3D voxels, which incur high communication costs, or 2D planar features, which require accurate depth estimation or additional supervision, limiting their applicability to collaborative scenarios. To address these challenges, we propose the first approach leveraging sparse 3D semantic Gaussian splatting for collaborative 3D semantic occupancy prediction. By sharing and fusing intermediate Gaussian primitives, our method provides three benefits: a neighborhood-based cross-agent fusion that removes duplicates and suppresses noisy or inconsistent Gaussians; a joint encoding of geometry and semantics in each primitive, which reduces reliance on depth supervision and allows simple rigid alignment; and sparse, object-centric messages that preserve structural information while reducing communication volume. Extensive experiments demonstrate that our approach outperforms single-agent perception and baseline collaborative methods by +8.42 and +3.28 points in mIoU, and +5.11 and +22.41 points in IoU, respectively. When further reducing the number of transmitted Gaussians, our method still achieves a +1.9 improvement in mIoU, using only 34.6\% communication volume, highlighting robust performance under limited communication budgets.
\end{abstract}

\begin{links}
    \link{Code}{https://github.com/ChengChen2020/VOGS-CP}
\end{links}
\begin{figure}[t]
    \centering
    \includegraphics[width=\columnwidth]{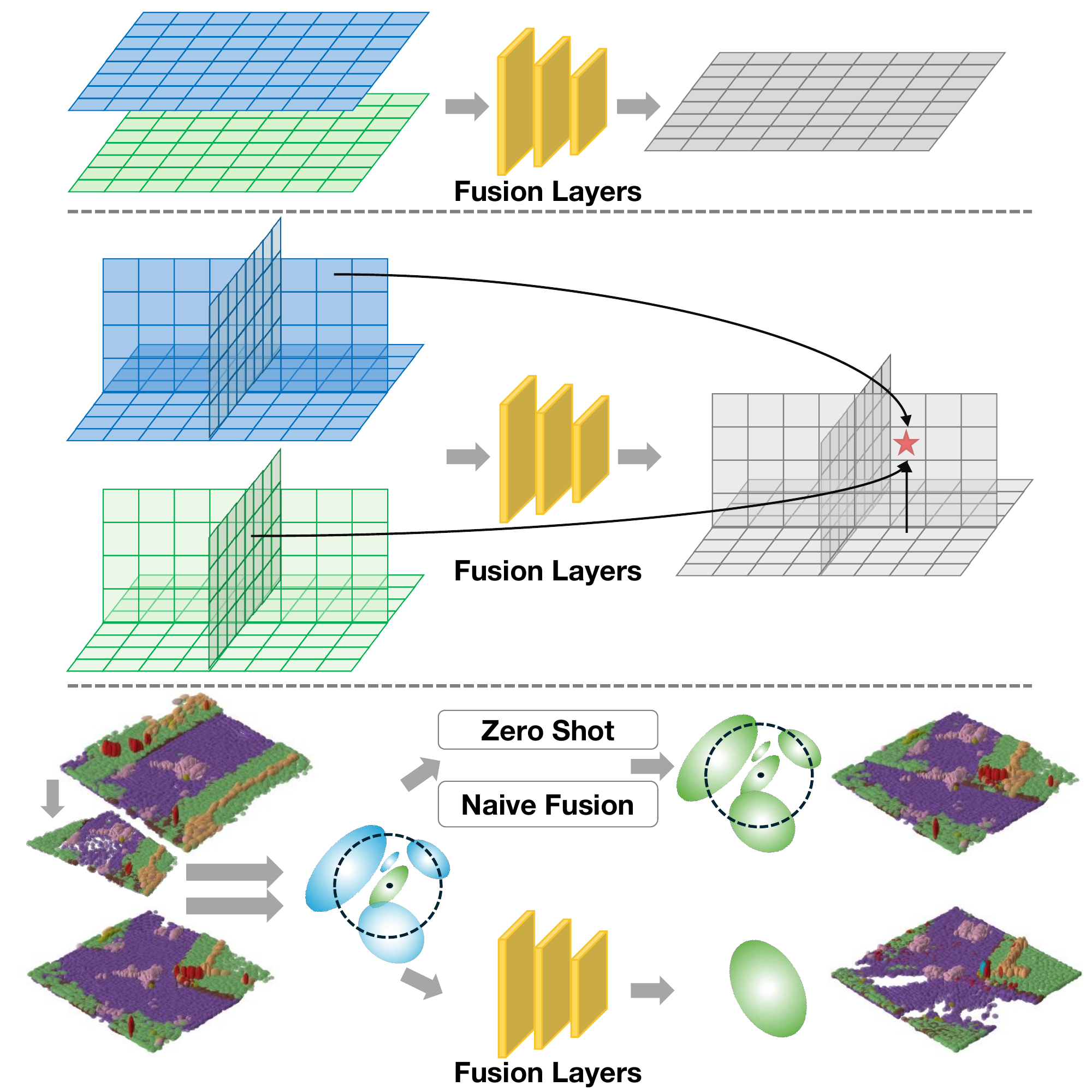}
    \caption{Comparison of shared representations. BEV and tri-plane methods transmit implicit planar features as messages, losing geometric detail and complicating alignment. We instead share explicit, interpretable 3D Gaussian primitives that preserve 3D structure and enable straightforward cross-agent fusion.}
    \label{fig:Communication_medium}
\end{figure}

\section{Introduction}

Multi-agent collaborative perception (CP), also known as cooperative perception, significantly enhances transportation safety, mobility, and efficiency by enabling connected vehicles to integrate multiple viewpoints through Vehicle-to-Everything (V2X) communication, forming a comprehensive understanding of their environment~\cite{cp_survey}. Early CP studies primarily focused on established single-agent tasks such as 3D object detection (3DOD)~\cite{who2com,where2comm,cppc,ermvp,v2xvit,v2v4real} and 2D bird's-eye-view (BEV) semantic segmentation~\cite{v2vnet,disconet}, typically by fusing latent BEV features (Figure~\ref{fig:Communication_medium} top). Historically reliant on LiDAR point clouds for their precise geometric measurements, recent advances in vision-only methods have shown competitive performance. For example, CoCa3D~\cite{coca3d} demonstrates that camera-based 3DOD can match or surpass LiDAR-based approaches, highlighting cameras as viable, cost-effective sensors for complex 3D scene understanding.

However, tasks like 3DOD and BEV segmentation simplify the perception of the scene by omitting crucial 3D semantic details, and BEV-based features inherently suffer from information loss due to height compression. This limitation hinders holistic scene understanding and downstream decision-making. 3D semantic occupancy prediction (SOP)~\cite{ssc-survey,occ3d,sscbench} bridges this gap by predicting the occupancy status of each voxel in the surrounding 3D space, thus delivering fine-grained semantic and geometric scene information. Vision-centric SOP methods~\cite{monoscene,ssc1,ssc2,voxformer,surroundocc,tpvformer} have shown promising results on large-scale datasets, yet extending SOP to collaborative scenarios remains largely unexplored. CoHFF~\cite{cohff} introduced the first collaborative SOP framework, demonstrating the benefits of V2X feature fusion. Nevertheless, it relies on planar tri-plane features~\cite{tpvformer,kplanes} (Figure~\ref{fig:Communication_medium} middle) which requires explicit depth supervision, and employs complex multi-stage, multi-task training pipelines, which introduces computational inefficiencies and complicates cross-agent feature alignment. Inspired by advances in 3D Gaussian splatting~\cite{3dgs}, sparse, object-centric representations have emerged as promising alternatives to traditional dense voxel or planar encodings for SOP~\cite{voxformer,surroundocc,tpvformer}. Driving scenes are highly sparse—most voxels are empty—so anisotropic Gaussians can shape and orient along semantic object surfaces, concentrating capacity where geometry and semantics lie while keeping free space compact, yielding an efficient 3D scene model. GaussianFormer~\cite{huang2024gaussianformer,huang2024gsformer2} represents scenes using sets of 3D Gaussians, each defined by a mean, covariance, and semantic label; these Gaussians are refined via deformable attention and rendered to a voxel grid through Gaussian-to-voxel splatting.

In this work, we propose the first Vision-Only Gaussian Splatting framework for collaborative SOP that can be efficiently trained end-to-end in a single stage. Our method leverages sparse 3D semantic Gaussians as intermediate representations shared among vehicles, explicitly encoding both geometric and semantic information. By transmitting rigid transforms of Gaussians only within each ego vehicle's region of interest (Figure~\ref{fig:Communication_medium} bottom), our approach effectively integrates multi-agent viewpoints, overcoming occlusions. Additionally, we introduce a novel cross-agent Gaussian fusion module that refines noisy, redundant, and conflicting Gaussians through neighborhood-based fusion, enabling accurate and realistic semantic occupancy prediction. To the best of our knowledge, we are the first to utilize 3D Gaussian splatting for multi-agent collaborative 3D semantic occupancy prediction. We summarize our contributions as:
\begin{itemize}[leftmargin=*]
\item We propose the first vision-only Gaussian splatting framework for collaborative SOP, which aggregates 3D Gaussians across agents, improving robustness to occlusions and viewpoint fragmentation.
\item We employ a learned neighborhood-based fusion module specifically designed to reduce noise and inconsistencies across multi-agent predictions.
\item Extensive experiments validate our framework's effectiveness and communication efficiency compared to single-agent and existing collaborative SOP methods, achieving robust performance in downstream tasks such as BEV semantic segmentation.
\end{itemize}
\section{Related Work}

\paragraph{3D Semantic Occupancy Prediction.}

Semantic occupancy prediction (SOP), also known as semantic scene completion (SSC), estimates per-voxel occupancy together with semantic categories, yielding a volumetric representation of geometry and semantics. Since Tesla described an occupancy network for Full Self-Driving~\cite{tesla}, research on 3D occupancy for autonomous driving has steadily accelerated. Methods differ by sensing modality: some use LiDAR point clouds~\cite{gaussianformer3d} or 4D radar tensors (4DRTs)~\cite{radarocc}, while many recent systems are vision-only and lift image features into 3D using learned geometric priors~\cite{monoscene,ssc1,ssc2,ssc3,tpvformer,huang2024gaussianformer,surroundocc,occformer,voxformer}. Most prior work addresses single-vehicle perception. Extending SOP to multi-agent collaborative perception centers on two design decisions: the communication representation and the fusion strategy for processing aggregated information under bandwidth limits and field-of-view differences. Collaborative SOP is only beginning to be explored; CoHFF~\cite{cohff} reports clear benefits from sharing but relies on multi-stage training and depth estimation, which increases system complexity and complicates cross-agent feature alignment.

\paragraph{Collaborative Perception.}

Collaborative perception for connected autonomous vehicles (CAVs) uses V2X communication and data fusion to improve scene understanding. Fusion is commonly grouped into \emph{early}, \emph{intermediate}, and \emph{late}~\cite{cp_survey}. Early fusion shares raw sensor data (e.g., point clouds), which is bandwidth demanding and raises privacy concerns~\cite{vieye}. Late fusion exchanges object lists, which is lightweight but discards fine details~\cite{vips,otvic}. Most recent systems adopt \emph{intermediate fusion}, which exchanges latent features and retains more task signal while avoiding raw data transfer~\cite{v2vnet,where2comm,cobevt,v2xvit,v2xvitv2,cppc}. While many works target LiDAR or hybrid setups for 3D detection or BEV segmentation~\cite{v2v4real,cppc,ermvp,tumtraf,disconet,where2comm,how2comm}, vision-only collaboration is increasingly practical for large-scale deployment. CoHFF brought \emph{collaborative semantic occupancy prediction} into focus by demonstrating SOP benefits from V2X feature sharing and fusion~\cite{cohff}.


\paragraph{3D Gaussians for Collaboration.}

3D Gaussian Splatting (3DGS) represents scenes with anisotropic Gaussians and renders them by splatting for novel view synthesis~\cite{3dgs}. Subsequent work adapts Gaussian splatting to perception tasks, including object detection~\cite{3dgs-det,gaussiandet}. For semantic occupancy, the GaussianFormer family splats Gaussians into voxels~\cite{huang2024gaussianformer,huang2024gsformer2}. Prior work treats Gaussians as an \emph{internal} single‑agent representation. We instead introduce Gaussians as the \emph{communication} medium for collaboration. This choice provides compact messages, closed‑form rigid alignment across agents, region‑of‑interest culling, and explicit geometry that reduces reliance on separate depth supervision. Our method builds communication and cross-agent fusion for Gaussian primitives.

\section{Proposed Approach}

\subsection{Problem Formulation}
\label{subsec:problem}
We first define a general collaborative perception pipeline at the feature level. Let $\mathcal{S}$ be the set of $N$ connected autonomous vehicles (CAVs) within a communication range $\delta$. For the $i$th agent in the set $\mathcal{S}$, we denote $\mathbf{O}_i$ as its observation, $f_\text{enc}(\cdot)$ as its perception encoder, $f_\text{fusion}(\cdot)$ as a fusion sub-network, $f_\text{head}(\cdot)$ as its task-specific head layers, and $\mathbf{B}_i$ as the corresponding task-specific output. Then, the collaborative perception network of the $i$th agent works as follows:
\begin{align}
    \mathbf{F}_i&=f_\text{enc}(\mathbf{O}_i),\enspace i\in\mathcal{S}\enspace,\\
    \mathbf{F}_{j\rightarrow i}&=\Gamma_{j\rightarrow i}(\mathbf{F}_j),\enspace j\in\mathcal{S}\enspace,\\
    \mathbf{H}_i&=f_\text{fusion}(\{\mathbf{F}_{j\rightarrow i}\}_{j\in\mathcal{S}})\enspace,\\
    \mathbf{B}_i&=f_\text{head}(\mathbf{H}_i)\enspace,
\end{align}
where $\mathbf{F}_i$ is the initial intermediate representation from the $i$th agent's encoder, $\Gamma_{j\rightarrow i}$ is an operator that performs spatial alignment and transmits $j$th agent's representation, $\mathbf{F}_{j\rightarrow i}$ is the spatially aligned representation of $j$th agent's observation in $i$th agent's coordinate frame, $\mathbf{H}_i$ is the fused aligned representations from the all neighbor agents in the set $\mathcal{S}$.

For vision-only semantic occupancy prediction, observation $\mathbf{O}_i$ is the RGB images captured by multiple surrounding cameras mounted on the $i$th agent, and $\mathbf{B}_i$ is the holistic surrounding environment represented as a 3D voxels with one-hot embedding, \ie, $\mathbf{B}_i\in\mathbb{R}^{X\times Y\times Z\times C}$ where $X,Y$ and $Z$ are voxel grid dimensions and $C$ denotes semantic classes. Let $\mathbf{\hat{B}}_i$ represent the ground-truth of the semantic voxels, and $\Phi$ represent the collaborative perception network (including the encoder, fusion sub-network, and head layers) parametrized by $\theta$, the objective of collaborative semantic occupancy prediction is defined as follows:
\begin{equation}
\small
    \max_{\theta}\sum_i g\left(\Phi_\theta(\mathbf{O}_i,\{\mathbf{F}_{j\rightarrow i}\}_{j\in\mathcal{S}}), \mathbf{\hat{B}}_i\right), \enspace \text{s.t.} \enspace \sum_j\left|\mathbf{F}_{j\rightarrow i}\right|\le \beta\enspace,
\end{equation}
where $g(\cdot,\cdot)$ is the metric for optimization. For the semantic occupancy prediction task, we adopt IoU and mIoU. The size of transmitted messages is constrained by a communication budget upper bound $\beta\in\mathbb{R}^+$.

Considering practical bandwidth limits, extending voxel-based single-agent SOP methods (e.g.,\cite{surroundocc,voxformer}) to collaboration is not viable: transmitting dense voxel features in $\mathbb{R}^{X\times Y\times Z\times D}$ (where $D$ is the hidden dimension) is too costly. Drawn inspiration from TPVFormer~\cite{tpvformer}, CoHFF proposes to transmit features from orthogonal planes $\mathbf{P}^{xz}$ and $\mathbf{P}^{yz}$, with features $\mathbf{F}^{\mathbf{P}^{xz}}\in \mathbb{R}^{X\times Z\times F}$ and $\mathbf{F}^{\mathbf{P}^{yz}}\in\mathbb{R}^{Y\times Z\times F}$. This reduces the communication volume from $XYZD$ to $(XZ+YZ)D$ and makes collaborative SOP feasible.

However, plane-based features (for example, BEV or tri-planes) do not encode explicit depth or full 3D geometry, so they need extra supervision. TPVFormer uses sparse semantic LiDAR labels, whereas CoHFF trains a separate depth estimation network. In addition, CoHFF adopts a two-stage schedule: the occupancy predictor and the semantic segmenter are trained first, followed by the fusion model. This increases training cost and reduces deployment scalability.


\subsection{Scene as 3D Gaussian Representation}

Inspired by GaussianFormer~\cite{huang2024gaussianformer}, we opt to represent a scene with a set of 3D Gaussian primitives $\mathcal{G}=\{\mathbf{G}_i\in\mathbb{R}^d|i=1,\ldots,P\}$. Distinct from GaussianFormer, we make the set of 3D Gaussian primitives serve as a communication medium for collaborative perception. Specifically, each $\mathbf{G}_i$ describes a local region with its mean $\mathbf{m}_i\in \mathbb{R}^3$, scale $\mathbf{s}_i\in\mathbb{R}^3$, rotation $\mathbf{r}_i\in\mathbb{R}^4$, opacity $a_i\in\mathbb{R}^1$, and semantics $\mathbf{c}_i\in\mathbb{R}^{|\mathcal{C}|}$. These primitives are interpreted as local semantic Gaussian distributions which contribute to the overall occupancy
prediction for any given 3D location $\mathbf{x} \in \mathbb{R}^3$ through additive aggregation:
\begin{align}
    \hat{\mathbf{o}}(\mathbf{x};\mathcal{G})=\sum_{i=1}^P\mathbf{g}_i(\mathbf{x};\mathbf{m}_i,\mathbf{s}_i,\mathbf{r}_i,a_i,\mathbf{c}_i)\enspace,
    \label{eq:additive}
\end{align}
where $\mathbf{g}_i(\mathbf{x};\cdot)$ denotes the contribution of the $i$th semantic Gaussian primitive to $\hat{\mathbf{o}}(\mathbf{x};\mathcal{G})$:
\begin{align}
    \mathbf{g}(\mathbf{x};\mathbf{G})=a\cdot\exp\left(-\frac{1}{2}(\mathbf{x}-\mathbf{m})^\text{T}\mathbf\Sigma^{-1}(\mathbf{x}-\mathbf{m})\right)\mathbf{c}\enspace,
\end{align}
\begin{align}
    \mathbf{\Sigma}=\mathbf{RSS}^\top\mathbf{R}^\top\enspace, \enspace \mathbf{S}=\text{diag}(\mathbf{s})\enspace, \enspace \mathbf{R}=\text{q2r}(\mathbf{r})\enspace,
\end{align}
where $\mathbf{\Sigma}$, $\mathbf{R}$, $\mathbf{S}$ represent the covariance matrix, the rotation matrix constructed from the quaternion $\mathbf{r}$ with function $\text{q2r}(\cdot)$, and the diagonal scale matrix from function $\text{diag}(\cdot)$.

\begin{figure*}[t]
    \centering
    \includegraphics[width=0.92\linewidth]{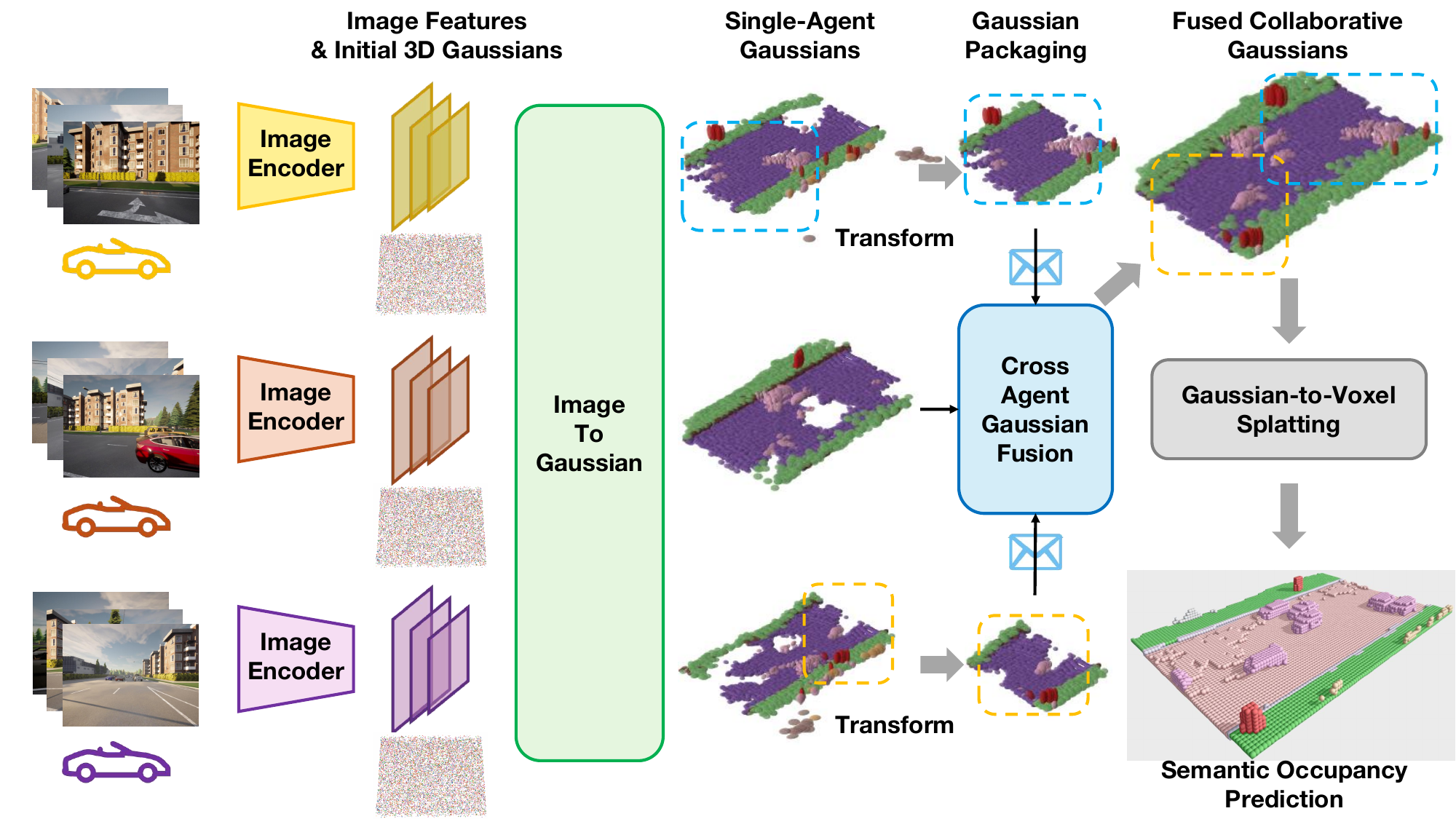}
    \caption{Overview of the proposed pipeline. An initial set of randomly initialized 3D Gaussians is refined by an Image-to-Gaussian module~\cite{huang2024gaussianformer} that attends to multi-scale image features, producing single-agent Gaussians. Neighbor agents (top and bottom) are rigidly transformed into the ego frame (middle) and culled to the ego region of interest; the blue and yellow dashed box marks the Gaussians that lie within the ego ROI and are packaged and transmitted to the ego. A cross-agent Gaussian fusion module aggregates these with the ego set. The fused Gaussians are then rendered to semantic occupancy via Gaussian-to-voxel splatting~\cite{huang2024gaussianformer}. For clarity, the figure shows the \emph{zero-shot} variant.}
    \label{fig:main_framework}
\end{figure*}

From a single-agent perspective, the pipeline first estimates 3D Gaussian primitives with an image-to-Gaussian module, then renders semantic occupancy via Gaussian-to-voxel splatting. We adopt both components from GaussianFormer~\cite{huang2024gaussianformer}.
We use 3D Gaussian primitives as a compact communication interface for collaborative perception. A \emph{Gaussian packaging} module transforms each neighbor’s Gaussians into the ego frame and culls them to the ego region of interest, reducing the transmitted data. A \emph{cross-agent Gaussian fusion} module then refines the ego set by aggregating consistent multi-view evidence and down-weighting noisy or low-opacity inputs. Collaborative semantic occupancy is obtained by splatting the refined ego Gaussians. The system is illustrated in Fig.~\ref{fig:main_framework}.

\subsection{Gaussian Packaging}
\label{sec:transform-transmit}

Unlike dense voxel grid features~\cite{surroundocc,voxformer} or planar features~\cite{cohff,tpvformer}, Gaussian primitives model ellipsoidal probability densities in $\mathbb{R}^3$, which remain well defined under rigid transforms~\cite{3dgs,huang2024gaussianformer}, and thus cross-agent 3D Gaussian alignment and transmission are straightforward and interpretable. Specifically, let the known extrinsic from agent \(j\) to agent \(i\) be
\(
T_{ij}(x)=U_{ij}x+t_{ij}
\)
with \(U_{ij}\in\mathrm{SO}(3)\), \(t_{ij}\in\mathbb{R}^3\), and quaternion \(q_{ij}\) such that \(\text{q2r}(q_{ij})=U_{ij}\).

\paragraph{Rigid Transform of a Gaussian.}
If \(X\!\sim\!\mathcal{N}(\mathbf{m},\mathbf{\Sigma)}\) in \(j\)th coordinate frame, then \(Y=T_{ij}(X)\) is Gaussian in \(i\)'th with:
\begin{align}
\mathbf{m}' &= U_{ij}\,\mathbf{m} + t_{ij}\enspace, 
\label{eq:mean-xf}\\
\mathbf{\Sigma}' &= U_{ij}\,\mathbf{\Sigma}\,U_{ij}^\top
       \;=\; \big(U_{ij}\mathbf{R}\big)\,\mathbf{S}\mathbf{S}^\top\,\big(U_{ij}\mathbf{R}\big)^\top\enspace. \label{eq:cov-xf}
\end{align}
Equations \eqref{eq:mean-xf}–\eqref{eq:cov-xf} imply that a rigid transform rotates the ellipsoid but does not change its axis lengths. In \((\mathbf{m},\mathbf{s},\mathbf{r},a,\mathbf{c})\) form, we have:
\begin{align}
\mathbf{m}' = U_{ij}\mathbf{m}+t_{ij}\enspace,\enspace
\mathbf{s}' = \mathbf{s}\enspace,\enspace \notag\\
\mathbf{r}' = q_{ij}\otimes \mathbf{r}\enspace,\enspace
a' = a\enspace,\enspace
\mathbf{c}' = \mathbf{c}\enspace,
\end{align}
where \(\otimes\) is quaternion multiplication. Because \(r\) and \(-r\) encode the same rotation, one may fix a sign convention (\eg, non-negative scalar part) without affecting the transform.

\paragraph{Transmission of a Gaussian.}
The $j$th agent only transmits Gaussians whose \emph{transformed} means fall inside the $i$th agent's region of interest \(\mathbf{ROI}_i\) (for example, a 3D volume centered at agent $i$):
%
\begin{align}
\mathcal{G}_{j\rightarrow i}
\;=\;
\{\,\mathbf{G}\in\mathcal{G}_j \;|\; \mathbf{m}' \in \mathbf{ROI}_i \,\}\enspace.
\end{align}
which greatly reduces the payload compared with transmitting all Gaussians \(|\mathcal{G}_j|\) from the $j$th agent. On reception, the $i$th agent stacks both its own and the received Gaussians in the same coordinate frame:
\begin{align}
\mathcal{G}^{\mathrm{stack}}_i
\;=\;
\mathcal{G}_i
\;\cup\;
\{\mathcal{G}_{j\rightarrow i}\}_{j\in\mathcal{S}}\enspace,
\end{align}
and proceeds with downstream processing.

Even without joint training, \emph{zero-shot} stacking—using single-agent weights—improves collaborative perception; collective end-to-end training yields further gains (see \emph{Zero Shot} and \emph{Naive Fusion} in Table~\ref{tab:main}), demonstrating the benefits of an explicit, interpretable representation.

\subsection{Cross-Agent Gaussian Fusion}
\label{sec:fusion}

Despite the fact that stacking transformed Gaussians improves the performance of collaborative perception for CAVs in semantic occupancy prediction, single agent predictions could be noisy due to occlusions, and different agents may output redundant or inconsistent primitives (Figure~\ref{fig:visualization} last row). We therefore propose a light-weight learnable refinement module that update the ego Gaussians via neighborhood interaction and aggregation.

Unlike the refinement step in the image-to-Gaussian module of GaussianFormer~\cite{huang2024gaussianformer}, our fusion module decodes neighborhood-conditioned proposals, pools them across nearby Gaussians to suppress noise and enforce consistency, and then blends the pooled update with the ego Gaussian attributes.
\begin{table*}[htbp]
    \centering
    \small
    \setlength{\tabcolsep}{9pt}
    \begin{tabular}{r| cc| c ccc}
        \toprule
\multirow{2}{*}{Method} &
\multicolumn{2}{c|}{Single-Agent} &
\multicolumn{4}{c}{Collaborative Perception} \\
\cmidrule(lr){2-3}\cmidrule(lr){4-7}
& CoHFF & GSFormer & CoHFF & Zero Shot & Naive Fusion & Learned Fusion \\
\midrule
        \midrule
        IoU $(\uparrow)$        & 38.52  & \textbf{67.76} & 50.46 & 67.88 & \underline{70.10} & \textbf{72.87}\\
        mIoU $(\uparrow)$        & 24.81  & \textbf{29.20} & 34.16 & 30.54 & \underline{36.02} & \textbf{37.44}\\
        \midrule
        Building    & \textbf{21.04}  & 3.84 & \textbf{25.72}  & 3.91  & 7.18 & \underline{9.61} \\
        Fence       & \textbf{20.05}  & 14.10 & \underline{27.83} & 17.29 & 25.66 & \textbf{29.20} \\
        Terrain     & 43.93  & \textbf{68.97} & 48.30 & 72.00 & \textbf{76.05} & \underline{74.51}  \\
        Pole        & \textbf{31.66}  & 5.94 & \textbf{42.74}  & 8.44  & \underline{12.67} & 12.19  \\
        Road        & 55.83 & \textbf{79.37} & 61.77 & \underline{81.35} & 78.60 & \textbf{83.05}  \\
        Side walk   & 17.31 & \textbf{70.55} & 39.62 & 68.91 & \underline{75.88} & \textbf{78.22}  \\
        Vegetation  & \textbf{14.49}  & 12.54 & \textbf{20.59} & 13.72 & 16.42 & \underline{20.43}  \\
        Vehicles    & \textbf{58.55}  & 49.25 & \textbf{63.28} & 49.56 & 57.14 & \underline{60.49}  \\
        Wall        & \textbf{33.30}  & 30.79 & \textbf{58.27} & 30.49 & 35.25 & \underline{36.45}  \\
        Guard rail  & 1.54  & \textbf{15.01} & 1.94  & 20.78 & \textbf{41.00} & \underline{32.50}  \\
        Traffic signs & 0    & 0 & \textbf{16.33}     & 0     & 6.35 & \underline{8.26}  \\
        Bridge      & 0     & 0   & \underline{3.53}    & 0     & 0 & \textbf{4.35}  \\
        \bottomrule
    \end{tabular}
    \caption{Comparison for semantic occupancy prediction. In the collaborative setting, our \emph{naive fusion} and \emph{learned fusion} variants achieve better IoU and mIoU than CoHFF~\cite{cohff}. The best is in \textbf{bold} and the second best is \underline{underlined}.}
    \label{tab:main}
\end{table*}

\paragraph{Neighborhood and Pairwise Features.}
Let $\mathcal{G}_i=\{\mathbf{G}_k\}_{k=1}^{N_i}$ be the ego Gaussians generated by the $i$th agent and $\mathcal{G}_{\text{nbgr}}=\bigcup_{j}\mathcal{G}_{j\rightarrow i}$ be received Gaussians (both in $i$th's coordinate fraome).
For each ego $\mathbf{G}_k=(\mathbf{m}_k,\mathbf{s}_k,\mathbf{r}_k,a_k,\mathbf{c}_k)$ we form a radius-$\rho$ neighborhood:
\begin{equation}
\mathcal{G}_{\text{nbgr},k}=\{\,\mathbf{G}_j\in\mathcal{G}_{\text{nbgr}}\mid \|\mathbf{m}_j-\mathbf{m}_k\|_2\le\rho\,\}\enspace.
\end{equation}
For each pair of Gaussians $\mathbf{G}_k \in \mathcal{G}_i$ and $\mathbf{G}_{k,j} \in \mathcal{G}_{\text{ngbr},k}$, we build a pairwise feature by concatenating the ego attributes with neighboring cues:
\begin{equation}
\begin{aligned}
\mathbf{f}^{\mathrm{ego}}_k &= \big[\,\mathbf{m}_k^\top,\ \mathbf{s}_k^\top,\ \mathbf{r}_k^\top,\ a_k,\ \mathbf{c}_k^\top\,\big]\enspace,\\
\mathbf{f}^{\mathrm{rel}}_{k,j} &= \big[\, (\mathbf{m}_j{-}\mathbf{m}_k)^\top,\ (\mathbf{s}_j{-}\mathbf{s}_k)^\top,\ |\langle \mathbf{r}_j,\mathbf{r}_k\rangle|,\ a_j,\ \mathbf{c}_j^\top\,\big]\enspace,\\
\mathbf{z}_{k,j} &= \big[\,\mathbf{f}^{\mathrm{ego}}_k \ \big| \ \mathbf{f}^{\mathrm{rel}}_{k,j}\,\big]\enspace, 
\end{aligned}
\end{equation}
where the term $|\langle \mathbf{r}_j,\mathbf{r}_k\rangle|$ is the sign-invariant quaternion cosine; opacity $a_j$ and semantics $\mathbf{c}_j$ are used in absolute form.

\paragraph{Pairwise Feature Refinement.}
A multi-layer perceptron (MLP) maps $\mathbf{z}_{k,j}$ to a refinement proposal from neighbor $j$ on ego Gaussian $k$:
%
\begin{equation}
\mathbf{u}_{k,j}\coloneq\big(\Delta \mathbf{m}_{k,j},\ \mathbf{s}^\star_{k,j},\ \mathbf{r}^\star_{k,j},\ a^\star_{k,j},\ \mathbf{c}^\star_{k,j}\big)
\;=\; \text{MLP}(\mathbf{z}_{k,j})\enspace,
\end{equation}
which are, respectively, a residual update to the Gaussian center, a scale proposal, a quaternion rotation proposal, an opacity proposal, and semantic logits.

\paragraph{Aggregation over Neighbors.}
We aggregrate pairwise proposals with either mean pooling or attention pooling.
Mean pooling uses uniform weights $w_{k,j}=1/|\mathcal{G}_{\text{nbgr},k}|$.
Attention pooling uses a learned softmax over neighbors:
\begin{equation}
w_{k,j}=\mathrm{softmax}_{j}\!\left(\frac{\langle Q\,\mathbf{x}^{\text{ego}}_k,\ K\,\mathbf{x}^{\text{rel}}_{k,j}\rangle}{\sqrt{d}}\right)\enspace,
\end{equation}
where $Q,K$ are learned projection layers. The aggregated proposal is:
\begin{equation}
\bar{\mathbf{u}}_k=\sum_{G_j\in\mathcal{G}_{\text{nbgr},k}} w_{k,j}\,\mathbf{u}_{k,j}\enspace.
\end{equation}

\paragraph{Update Rules.}
Let $\bar{\mathbf{u}}_k=(\overline{\Delta \mathbf{m}}_k,\overline{\mathbf{s}^\star}_k,\overline{\mathbf{r}^\star}_k,\overline{a^\star}_k,\overline{\mathbf{c}^\star}_k)$.
We update the ego Gaussian as:
\begin{equation}
\begin{aligned}
\hat{ \mathbf{m}_k} &= \mathbf{m}_k + \overline{\Delta \mathbf{m}}_k\enspace, \qquad
\hat{ \mathbf{s}_k} \;=\; \overline{\mathbf{s}^\star}_k\enspace,\\
\hat{ \mathbf{r}_k} &= \overline{\mathbf{r}^\star}_k\enspace, \qquad
\hat a_k \;=\; \overline{a^\star}_k\enspace,\\
\hat{ \mathbf{c}_k} &= \alpha_k\,\mathbf{c}_k + (1{-}\alpha_k)\,\overline{\mathbf{c}^\star}_k\enspace,
\end{aligned}
\end{equation}
where the semantic blend uses a confidence weight:
\begin{equation}
\small
\alpha_k=\frac{\mathrm{conf}(\mathbf{c}_k)}{\mathrm{conf}(\mathbf{c}_k)+\mathrm{conf}(\overline{\mathbf{c}^\star}_k)}\,\enspace,
\quad
\mathrm{conf}(v)=\max\!\Big(\frac{v}{\mathbf{1}^\top v+\varepsilon}\Big)\enspace.
\end{equation}
\section{Experiments}

\paragraph{Datasets.}
OPV2V is a large-scale collaborative perception dataset collected in the CARLA simulator~\cite{dosovitskiy2017carla} using the OpenCDA autonomous driving simulation framework with Vehicle-to-Vehicle (V2V) communication, with each sample containing multi-sensor data from 2-7 vehicles. The original OPV2V does not provide semantic occupancy labels, so we use the enhanced Semantic-OPV2V released by CoHFF~\cite{cohff}, which replays the simulations with additional semantic LiDARs. Following the procedure in CoHFF, we aggregate the multi‑agent ground truth to obtain the collaborative semantic occupancy supervision.

\begin{table}[t]
    \centering
    \small
    \begin{tabular}{lcccc}
        \toprule
        Approach & \# Agents & Vehicle & Road & Others \\
        \midrule
        \midrule 
        CoBEVT & 2        & 46.13 & 52.41 & - \\
        CoHFF            & 2        & 47.40 & 63.36 & 40.27 \\
        Ours & 2 & \textbf{70.25} & \textbf{82.69} & \textbf{79.37} \\
        \midrule
        CoBEVT & Up to 7  & 60.40 & 63.00 & - \\
        CoHFF            & Up to 7  & 64.44 & 57.28 & 45.89 \\
        Ours             & Up to 7  & \textbf{75.30} & \textbf{84.96} & \textbf{80.19} \\
        \bottomrule
    \end{tabular}
    \caption{Comparison of BEV semantic segmentation with IoU (\%) for Vehicle, Road, and Others classes.}
    \label{tab:bev_semantic_iou}
\end{table}

\begin{table}[t]
\centering
\small
\setlength{\tabcolsep}{6pt}
\begin{tabular}{r|c|cc}
\toprule
\multirow{2}{*}{Metric} & \multirow{2}{*}{CoHFF} & \multicolumn{2}{c}{ \# Gaussians} \\
\cmidrule(lr){3-4}
 & & 25600 &  6400 \\
\midrule
CV (MB) ($\downarrow$) & 0.78 & 1.07  & 0.27 \\
IoU ($\uparrow$)       & 50.46 & 72.87  & 72.42 \\
mIoU ($\uparrow$)      & 34.16 & 37.44  & 36.02 \\
\bottomrule
\end{tabular}
    \caption{Comparison of communication volume.}
    \label{tab:communication_volume}
\end{table}

\paragraph{Implementation Details.}
Following CoHFF~\cite{cohff}, we utilize a $40\times40\times3.2$ meter detection area with a grid size of $100\times100\times8$, resulting in a voxel size of $0.4 m^3$. Unless otherwise noted, each agent default utilizes $|\mathcal{G}|=P=25600$ Gaussians as scene representation, and neighborhood $\rho$ is set to $0.4m$ with attention pooling. We allow CAVs to transmit and share Gaussian primitives for cross-agent fusion. Our experiment incorporates the analysis of 12 semantic labels plus an additional empty label. For optimization, we use AdamW~\cite{adamw} with weight decay $0.01$. The learning rate warms up for the first 500 iterations to $2\times10^{-4}$ and then follows a cosine decay. We train for 30 epochs with batch size 8 on a single NVIDIA A100 (40GB). Unless noted otherwise, all components are trained end to end. Following GaussianFormer~\cite{huang2024gaussianformer}, we use voxel‑wise cross‑entropy loss together with Lovász‑Softmax~\cite{berman2018lovasz} loss, which directly targets IoU improvement. The total loss is:
\begin{equation}
    L \;=\; L_{\mathrm{CE}} \;+\; L_{\mathrm{Lovasz}}\enspace.
\end{equation}

\paragraph{Evaluation Metrics.} Following~\cite{monoscene,tpvformer,cohff}, we adopt mean Intersection-over-Union (mIoU) and Intersection-over-Union (IoU) to evaluate the performance of semantic occupancy prediction.
%
For evaluations in subsequent applications, following prior work~\cite{cohff}, we report
BEV segmentation using 2D IoU by
projecting predicted and ground-truth semantic voxels onto the BEV plane and compute IoU for vehicles, roads and other general objects.


\subsection{Results and Analysis}
\label{sec:results}

\begin{table}[t]
\centering
\small
\setlength{\tabcolsep}{6pt}
\begin{tabular}{c c c | c c}
\toprule
\# Gaussians & Radius $\rho$ & Attn. & mIoU $\uparrow$ & IoU $\uparrow$ \\
\midrule
6400  & 0.4 &            & 35.50 & 71.81 \\
6400  & 0.8 &            & 36.02 & 72.42 \\
\midrule
25600 & 0.4 &            & 37.01 & 73.49 \\
25600 & 0.4 & \checkmark & 37.44 & 72.87 \\
25600 & 0.8 &            & 36.81 & 72.28 \\
25600 & 0.8 & \checkmark & 37.06 & 73.03 \\
\bottomrule
\end{tabular}
\caption{Ablation over number of Gaussians, neighborhood radius $\rho$, and pooling method.}
\label{tab:ablation_gauss_radius_attention}
\end{table}

\begin{figure*}[t]
\centering
    \includegraphics[width=\linewidth]{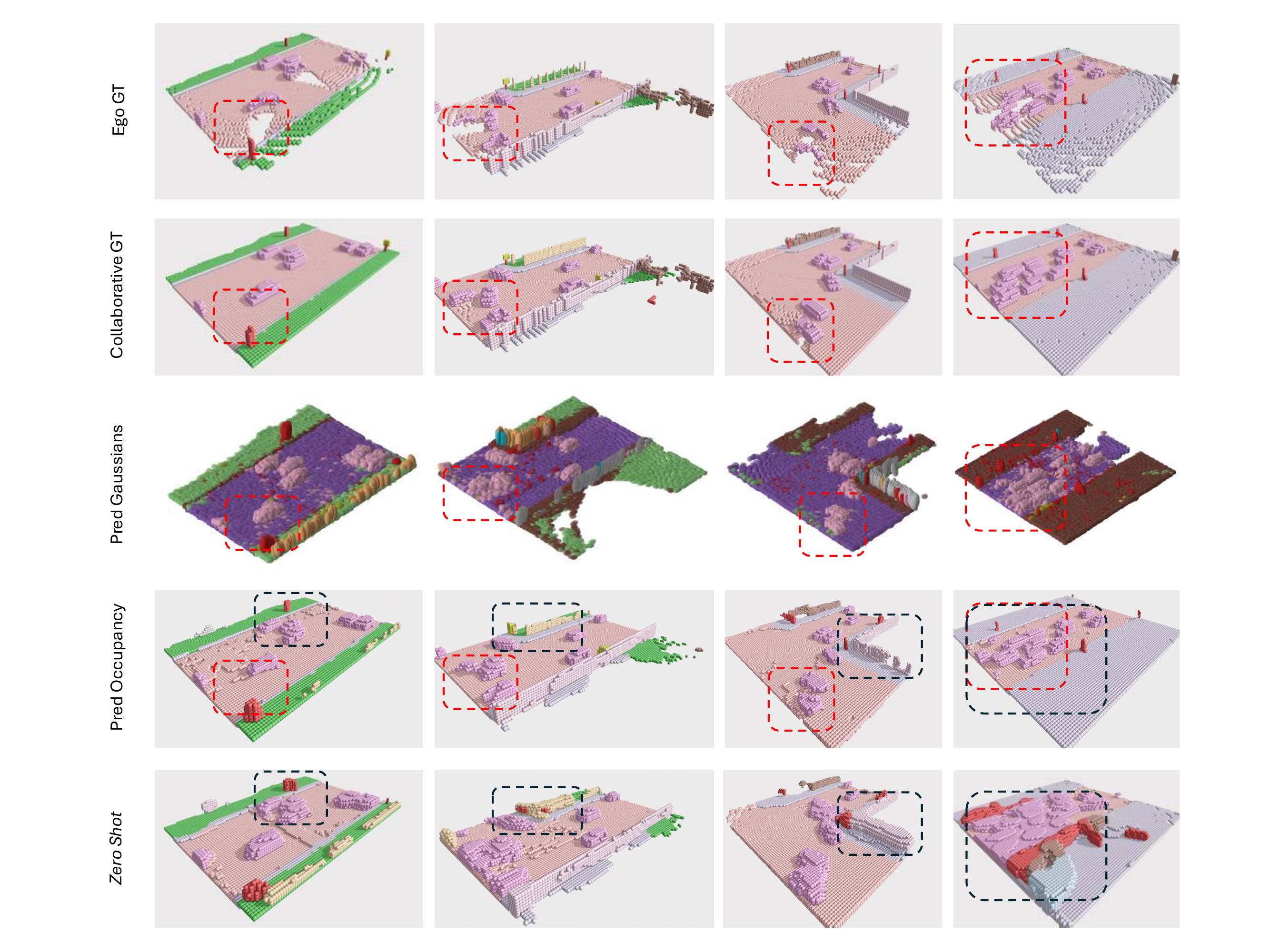}
    \caption{Qualitative comparison of ego-only ground truth, collaborative ground truth, predicted Gaussians, predicted occupancy, and the \emph{zero-shot} variant. Red boxes highlight occluded object structure captured by Gaussian primitives in collaborative setting. The \emph{zero-shot} variant can look plausible but often shows clustered redundancy and noise; black boxes mark cases where the neighborhood-based fusion suppresses redundancy and improves consistency. An opacity threshold is applied for display, so the predicted Gaussians are not exhaustive.}
\label{fig:visualization}
\end{figure*}

Table~\ref{tab:main} compares semantic occupancy prediction performance across CoHFF and our fusion variants.

\subsubsection{Single-Agent Perception.} We observe a large IoU gain (+29.24) steming from explicit free-space modeling: one large fixed Gaussian represents empty space across the scene, while the remaining Gaussians model occupied regions. This reduces confusion between empty and occupied voxels and allows the network focus capacity on object surfaces. The model learns the empty-space primitive and the occupied classes jointly, rather than using a separate occupancy network as in CoHFF. We further observe an increase in mIoU, driven chiefly by better performance on ground-level categories such as sidewalk and terrian. 

\subsubsection{Collaborative Perception.} In collaborative scenarios we evaluate three variants of our approach. The \textbfit{zero-shot} variant enables the ego vehicle directly concatenate Gaussians received from its neighbors before occupancy rendering. The \textbfit{naive fusion} variant trains the entire multi-agent system end-to-end starting from the zero-shot baseline. The \textbfit{learned fusion} variant applies the proposed cross-agent Gaussian fusion pooling to combine neighboring Gaussians.

The \textbfit{zero-shot} variant improves on the single-agent baseline by a small margin, confirming that Gaussian messages convey useful, explicit 3D evidence. However, stacking Gaussians from multiple agents without coordination introduces noise and inconsistency (see Figure \ref{fig:visualization}), which leads to redundant and inaccurate occupancy predictions. When the system is trained end-to-end on top of this baseline (i.e., \textbfit{naive fusion}), the resulting fusion variant achieves larger gains and surpasses CoHFF in mIoU, with performance on more classes (e.g., fence, vegetation, vehicles) matching or exceeding that of CoHFF. The \textbfit{learned fusion} variant refines the ego Gaussians by pooling neighbouring Gaussians shared by other agents, leading to measurable accuracy gains across most categories. Notably, the bridge class—missed entirely by the naive-shot variant—improves from 0\% to 4.35\% IoU. Though the increase is modest, it shows that the learned fusion helps capture structures that are both sparse and difficult to observe from a single viewpoint.

\subsubsection{BEV Segmentation.} Table \ref{tab:bev_semantic_iou} illustrates improved IoU for road and other semantic categories. Although the 3D mIoU for vehicle is slightly lower than that of CoHFF, projecting the predictions to the 2D plane yields a better BEV IoU, indicating that our voxel occupancy provides more accurate height estimates when mapping to ground level.

\subsubsection{Communication Volume.}

Communication volume, measured by the size of transmitted messages, is critical for cooperative perception because deployed systems must operate under limited bandwidth. Our framework exchanges variable-length sets of Gaussian primitives instead of fixed-resolution feature maps, so the communication load rises with the number of Gaussians that cover the overlapping regions of interest among connected vehicles. We therefore report the average message size over the evaluation set. As shown in Table \ref{tab:communication_volume}, the transmitted volume is roughly proportional to the number of Gaussians. By reducing this count, our method attains better occupancy prediction than CoHFF while using only 34.6\% bandwidth.

\subsubsection{Ablation Study.}

Table \ref{tab:ablation_gauss_radius_attention} lists an ablation in which we vary three factors: (i) the number of 3-D Gaussians, (ii) the neighbourhood radius $\rho$, and (iii) the pooling rule (mean versus attention pooling). Increasing the Gaussian count leads to a rise in mIoU, reflecting the finer geometric detail that a denser set of primitives can encode. Consistently high IoU across all settings demonstrates the effectiveness of using separate Gaussians to model free and occupied space. Changing $\rho$ or switching between the two pooling rules produces only minor fluctuations, indicating that the method is robust to these settings within the tested range.

\subsubsection{Visualizations.}
Figure \ref{fig:visualization} presents qualitative results together with two reference annotations: (i) the ground truth restricted to the ego vehicle’s field of view (Ego GT) and (ii) the combined ground truth from all connected vehicles (Collaborative GT). After exchanging Gaussian primitives, our model reconstructs more complete instances of vehicles, road surfaces, terrain, walls, and traffic signs than those visible in Ego GT, filling regions that were initially occluded. In some scenarios our prediction is even smoother and more continuous than Collaborative GT. Compared with the \emph{zero-shot} variant, whose direct stacking of Gaussians produces cluttered occupancy maps (bottom row), the proposed cross-agent Gaussian fusion outputs clean and coherent representations of the scene.

\section{Conclusion}

In this work, we demonstrate that explicit 3D Gaussian primitives serve as a compact, interpretable medium for vision-only cooperative 3D semantic occupancy prediction. Our pipeline aligns neighbor Gaussians via rigid transform and region of interest filtering, then refines the ego set with a lightweight, neighborhood-based fusion module. Experiments on Semantic-OPV2V validate that both naive and learned fusion variants achieve notable IoU and mIoU enhancements over planar-feature methods. 

\section{Acknowledgments}
This material is based in part upon work supported by the National Science Foundation under the Center CHORUS with Grant Number CNS-2333487, Army Research Lab under Contract number W911NF-20-2-0026, and gift funding from Amazon AWS. Any opinions, findings, and conclusions or recommendations expressed in this material are those of the authors and do not necessarily reflect the views of the sponsors.

\bibliography{aaai2026}

@inproceedings{berman2018lovasz,
  title={The lov{\'a}sz-softmax loss: A tractable surrogate for the optimization of the intersection-over-union measure in neural networks},
  author={Berman, Maxim and Triki, Amal Rannen and Blaschko, Matthew B},
  booktitle={Proceedings of the IEEE conference on computer vision and pattern recognition},
  pages={4413--4421},
  year={2018}
}

@inproceedings{huang2024gaussianformer,
  title={Gaussianformer: Scene as gaussians for vision-based 3d semantic occupancy prediction},
  author={Huang, Yuanhui and Zheng, Wenzhao and Zhang, Yunpeng and Zhou, Jie and Lu, Jiwen},
  booktitle={European Conference on Computer Vision},
  pages={376--393},
  year={2024},
  organization={Springer}
}

@inproceedings{dosovitskiy2017carla,
  title={CARLA: An open urban driving simulator},
  author={Dosovitskiy, Alexey and Ros, German and Codevilla, Felipe and Lopez, Antonio and Koltun, Vladlen},
  booktitle={Conference on robot learning},
  pages={1--16},
  year={2017},
  organization={PMLR}
}

@article{huang2024gsformer2,
  title={Probabilistic Gaussian Superposition for Efficient 3D Occupancy Prediction},
  author={Huang, Yuanhui and Thammatadatrakoon, Amonnut and Zheng, Wenzhao and Zhang, Yunpeng and Du, Dalong and Lu, Jiwen},
  journal={arXiv preprint arXiv:2412.04384},
  year={2024}
}

@inproceedings{v2vnet,
  title={V2vnet: Vehicle-to-vehicle communication for joint perception and prediction},
  author={Wang, Tsun-Hsuan and Manivasagam, Sivabalan and Liang, Ming and Yang, Bin and Zeng, Wenyuan and Urtasun, Raquel},
  booktitle={European conference on computer vision},
  pages={605--621},
  year={2020},
  organization={Springer}
}

@inproceedings{who2com,
  title={Who2com: Collaborative perception via learnable handshake communication},
  author={Liu, Yen-Cheng and Tian, Junjiao and Ma, Chih-Yao and Glaser, Nathan and Kuo, Chia-Wen and Kira, Zsolt},
  booktitle={2020 IEEE International Conference on Robotics and Automation (ICRA)},
  pages={6876--6883},
  year={2020},
  organization={IEEE}
}

@article{where2comm,
  title={Where2comm: Communication-efficient collaborative perception via spatial confidence maps},
  author={Hu, Yue and Fang, Shaoheng and Lei, Zixing and Zhong, Yiqi and Chen, Siheng},
  journal={Advances in neural information processing systems},
  volume={35},
  pages={4874--4886},
  year={2022}
}

@inproceedings{cppc,
  title={Point cluster: A compact message unit for communication-efficient collaborative perception},
  author={Ding, Zihan and Fu, Jiahui and Liu, Si and Li, Hongyu and Chen, Siheng and Li, Hongsheng and Zhang, Shifeng and Zhou, Xu},
  booktitle={The Thirteenth International Conference on Learning Representations},
  year={2025}
}

@inproceedings{surroundocc,
  title={Surroundocc: Multi-camera 3d occupancy prediction for autonomous driving},
  author={Wei, Yi and Zhao, Linqing and Zheng, Wenzhao and Zhu, Zheng and Zhou, Jie and Lu, Jiwen},
  booktitle={Proceedings of the IEEE/CVF International Conference on Computer Vision},
  pages={21729--21740},
  year={2023}
}

@inproceedings{cohff,
  title={Collaborative semantic occupancy prediction with hybrid feature fusion in connected automated vehicles},
  author={Song, Rui and Liang, Chenwei and Cao, Hu and Yan, Zhiran and Zimmer, Walter and Gross, Markus and Festag, Andreas and Knoll, Alois},
  booktitle={Proceedings of the IEEE/CVF Conference on Computer Vision and Pattern Recognition},
  pages={17996--18006},
  year={2024}
}

@inproceedings{monoscene,
  title={Monoscene: Monocular 3d semantic scene completion},
  author={Cao, Anh-Quan and De Charette, Raoul},
  booktitle={Proceedings of the IEEE/CVF Conference on Computer Vision and Pattern Recognition},
  pages={3991--4001},
  year={2022}
}

@inproceedings{tpvformer,
  title={Tri-perspective view for vision-based 3d semantic occupancy prediction},
  author={Huang, Yuanhui and Zheng, Wenzhao and Zhang, Yunpeng and Zhou, Jie and Lu, Jiwen},
  booktitle={Proceedings of the IEEE/CVF conference on computer vision and pattern recognition},
  pages={9223--9232},
  year={2023}
}

@inproceedings{coca3d,
  title={Collaboration helps camera overtake lidar in 3d detection},
  author={Hu, Yue and Lu, Yifan and Xu, Runsheng and Xie, Weidi and Chen, Siheng and Wang, Yanfeng},
  booktitle={Proceedings of the IEEE/CVF Conference on Computer Vision and Pattern Recognition},
  pages={9243--9252},
  year={2023}
}

@article{adamw,
  title={Decoupled weight decay regularization},
  author={Loshchilov, Ilya and Hutter, Frank},
  journal={arXiv preprint arXiv:1711.05101},
  year={2017}
}

@article{3dgs-det,
  title={3dgs-det: Empower 3d gaussian splatting with boundary guidance and box-focused sampling for 3d object detection},
  author={Cao, Yang and Jv, Yuanliang and Xu, Dan},
  journal={arXiv preprint arXiv:2410.01647},
  year={2024}
}

@article{3dgs,
  title={3D Gaussian splatting for real-time radiance field rendering.},
  author={Kerbl, Bernhard and Kopanas, Georgios and Leimk{\"u}hler, Thomas and Drettakis, George},
  journal={ACM Trans. Graph.},
  volume={42},
  number={4},
  pages={139--1},
  year={2023}
}

@article{gaussiandet,
  title={Gaussian-det: Learning closed-surface gaussians for 3d object detection},
  author={Yan, Hongru and Zheng, Yu and Duan, Yueqi},
  journal={arXiv preprint arXiv:2410.01404},
  year={2024}
}

@misc{tesla,
  author       = {Tesla},
  title        = {Tesla {AI} Day 2022},
  year         = {2022},
  howpublished = {\url{https://www.youtube.com/watch?v=ODSJsviD_SU}},
  note         = {YouTube video, accessed 2025-07-27}
}

@article{gaussianformer3d,
  title={GaussianFormer3D: Multi-Modal Gaussian-based Semantic Occupancy Prediction with 3D Deformable Attention},
  author={Zhao, Lingjun and Wei, Sizhe and Hays, James and Gan, Lu},
  journal={arXiv preprint arXiv:2505.10685},
  year={2025}
}

@article{cp_survey,
  title={V2X cooperative perception for autonomous driving: Recent advances and challenges},
  author={Huang, Tao and Liu, Jianan and Zhou, Xi and Nguyen, Dinh C and Azghadi, Mostafa Rahimi and Xia, Yuxuan and Han, Qing-Long and Sun, Sumei},
  journal={arXiv preprint arXiv:2310.03525},
  year={2023}
}

@inproceedings{vieye,
  title={VI-eye: semantic-based 3D point cloud registration for infrastructure-assisted autonomous driving},
  author={He, Yuze and Ma, Li and Jiang, Zhehao and Tang, Yi and Xing, Guoliang},
  booktitle={Proceedings of the 27th Annual International Conference on Mobile Computing and Networking},
  pages={573--586},
  year={2021}
}

@inproceedings{otvic,
  title={OTVIC: A Dataset with Online Transmission for Vehicle-to-Infrastructure Cooperative 3D Object Detection},
  author={Zhu, He and Wang, Yunkai and Kong, Quyu and Wei, Yufei and Xia, Xunlong and Deng, Bing and Xiong, Rong and Wang, Yue},
  booktitle={2024 IEEE/RSJ International Conference on Intelligent Robots and Systems (IROS)},
  pages={10732--10739},
  year={2024},
  organization={IEEE}
}

@inproceedings{vips,
  title={VIPS: Real-time perception fusion for infrastructure-assisted autonomous driving},
  author={Shi, Shuyao and Cui, Jiahe and Jiang, Zhehao and Yan, Zhenyu and Xing, Guoliang and Niu, Jianwei and Ouyang, Zhenchao},
  booktitle={Proceedings of the 28th annual international conference on mobile computing and networking},
  pages={133--146},
  year={2022}
}

@inproceedings{v2xvit,
  title={V2x-vit: Vehicle-to-everything cooperative perception with vision transformer},
  author={Xu, Runsheng and Xiang, Hao and Tu, Zhengzhong and Xia, Xin and Yang, Ming-Hsuan and Ma, Jiaqi},
  booktitle={European conference on computer vision},
  pages={107--124},
  year={2022},
  organization={Springer}
}

@article{v2xvitv2,
  title={V2x-vitv2: Improved vision transformers for vehicle-to-everything cooperative perception},
  author={Xu, Runsheng and Chen, Chia-Ju and Tu, Zhengzhong and Yang, Ming-Hsuan},
  journal={IEEE transactions on pattern analysis and machine intelligence},
  year={2024},
  publisher={IEEE}
}

@inproceedings{tumtraf,
  title={Tumtraf v2x cooperative perception dataset},
  author={Zimmer, Walter and Wardana, Gerhard Arya and Sritharan, Suren and Zhou, Xingcheng and Song, Rui and Knoll, Alois C},
  booktitle={Proceedings of the IEEE/CVF conference on computer vision and pattern recognition},
  pages={22668--22677},
  year={2024}
}

@inproceedings{v2v4real,
  title={V2v4real: A real-world large-scale dataset for vehicle-to-vehicle cooperative perception},
  author={Xu, Runsheng and Xia, Xin and Li, Jinlong and Li, Hanzhao and Zhang, Shuo and Tu, Zhengzhong and Meng, Zonglin and Xiang, Hao and Dong, Xiaoyu and Song, Rui and others},
  booktitle={Proceedings of the IEEE/CVF conference on computer vision and pattern recognition},
  pages={13712--13722},
  year={2023}
}

@inproceedings{ermvp,
  title={Ermvp: Communication-efficient and collaboration-robust multi-vehicle perception in challenging environments},
  author={Zhang, Jingyu and Yang, Kun and Wang, Yilei and Wang, Hanqi and Sun, Peng and Song, Liang},
  booktitle={Proceedings of the IEEE/CVF Conference on Computer Vision and Pattern Recognition},
  pages={12575--12584},
  year={2024}
}

@article{cobevt,
  title={CoBEVT: Cooperative bird's eye view semantic segmentation with sparse transformers},
  author={Xu, Runsheng and Tu, Zhengzhong and Xiang, Hao and Shao, Wei and Zhou, Bolei and Ma, Jiaqi},
  journal={arXiv preprint arXiv:2207.02202},
  year={2022}
}

@article{disconet,
  title={Learning distilled collaboration graph for multi-agent perception},
  author={Li, Yiming and Ren, Shunli and Wu, Pengxiang and Chen, Siheng and Feng, Chen and Zhang, Wenjun},
  journal={Advances in Neural Information Processing Systems},
  volume={34},
  pages={29541--29552},
  year={2021}
}

@inproceedings{ssc1,
  title={Semantic scene completion from a single depth image},
  author={Song, Shuran and Yu, Fisher and Zeng, Andy and Chang, Angel X and Savva, Manolis and Funkhouser, Thomas},
  booktitle={Proceedings of the IEEE conference on computer vision and pattern recognition},
  pages={1746--1754},
  year={2017}
}

@inproceedings{ssc2,
  title={3d sketch-aware semantic scene completion via semi-supervised structure prior},
  author={Chen, Xiaokang and Lin, Kwan-Yee and Qian, Chen and Zeng, Gang and Li, Hongsheng},
  booktitle={Proceedings of the IEEE/CVF Conference on Computer Vision and Pattern Recognition},
  pages={4193--4202},
  year={2020}
}

@inproceedings{ssc3,
  title={Anisotropic convolutional networks for 3d semantic scene completion},
  author={Li, Jie and Han, Kai and Wang, Peng and Liu, Yu and Yuan, Xia},
  booktitle={Proceedings of the IEEE/CVF Conference on Computer Vision and Pattern Recognition},
  pages={3351--3359},
  year={2020}
}

@inproceedings{voxformer,
  title={Voxformer: Sparse voxel transformer for camera-based 3d semantic scene completion},
  author={Li, Yiming and Yu, Zhiding and Choy, Christopher and Xiao, Chaowei and Alvarez, Jose M and Fidler, Sanja and Feng, Chen and Anandkumar, Anima},
  booktitle={Proceedings of the IEEE/CVF conference on computer vision and pattern recognition},
  pages={9087--9098},
  year={2023}
}

@inproceedings{kplanes,
  title={K-planes: Explicit radiance fields in space, time, and appearance},
  author={Fridovich-Keil, Sara and Meanti, Giacomo and Warburg, Frederik Rahb{\ae}k and Recht, Benjamin and Kanazawa, Angjoo},
  booktitle={Proceedings of the IEEE/CVF Conference on Computer Vision and Pattern Recognition},
  pages={12479--12488},
  year={2023}
}

@article{ssc-survey,
  title={3D semantic scene completion: A survey},
  author={Roldao, Luis and De Charette, Raoul and Verroust-Blondet, Anne},
  journal={International Journal of Computer Vision},
  volume={130},
  number={8},
  pages={1978--2005},
  year={2022},
  publisher={Springer}
}

@inproceedings{sscbench,
  title={Sscbench: A large-scale 3d semantic scene completion benchmark for autonomous driving},
  author={Li, Yiming and Li, Sihang and Liu, Xinhao and Gong, Moonjun and Li, Kenan and Chen, Nuo and Wang, Zijun and Li, Zhiheng and Jiang, Tao and Yu, Fisher and others},
  booktitle={2024 IEEE/RSJ International Conference on Intelligent Robots and Systems (IROS)},
  pages={13333--13340},
  year={2024},
  organization={IEEE}
}

@article{radarocc,
  title={Radarocc: Robust 3d occupancy prediction with 4d imaging radar},
  author={Ding, Fangqiang and Wen, Xiangyu and Zhu, Yunzhou and Li, Yiming and Lu, Chris Xiaoxuan},
  journal={Advances in Neural Information Processing Systems},
  volume={37},
  pages={101589--101617},
  year={2024}
}

@inproceedings{occformer,
  title={Occformer: Dual-path transformer for vision-based 3d semantic occupancy prediction},
  author={Zhang, Yunpeng and Zhu, Zheng and Du, Dalong},
  booktitle={Proceedings of the IEEE/CVF International Conference on Computer Vision},
  pages={9433--9443},
  year={2023}
}

@article{how2comm,
  title={How2comm: Communication-efficient and collaboration-pragmatic multi-agent perception},
  author={Yang, Dingkang and Yang, Kun and Wang, Yuzheng and Liu, Jing and Xu, Zhi and Yin, Rongbin and Zhai, Peng and Zhang, Lihua},
  journal={Advances in Neural Information Processing Systems},
  volume={36},
  pages={25151--25164},
  year={2023}
}

@article{occ3d,
  title={Occ3d: A large-scale 3d occupancy prediction benchmark for autonomous driving},
  author={Tian, Xiaoyu and Jiang, Tao and Yun, Longfei and Mao, Yucheng and Yang, Huitong and Wang, Yue and Wang, Yilun and Zhao, Hang},
  journal={Advances in Neural Information Processing Systems},
  volume={36},
  pages={64318--64330},
  year={2023}
}

\end{document}